\documentclass[conference]{IEEEtran}
\IEEEoverridecommandlockouts

\usepackage{cite}
\usepackage{amsmath,amssymb,amsfonts}
\usepackage{algorithmic}
\usepackage{graphicx}
\usepackage{textcomp}
\usepackage{xcolor}

\usepackage{array}
\usepackage{listings}
\usepackage{xcolor}

\definecolor{background}{RGB}{255,255,255}
\definecolor{numb}{RGB}{125,125,125}

\lstdefinelanguage{JSON}{
    string=[s]{"}{"},
    stringstyle=\color{red},
    numbers=none,
    numberstyle=\small,
    stepnumber=1,
    numbersep=8pt,
    showstringspaces=false,
    breaklines=true,
    frame=lines,
    backgroundcolor=\color{background},
    literate=
     *{0}{{{\color{numb}0}}}{1}
      {1}{{{\color{numb}1}}}{1}
      {2}{{{\color{numb}2}}}{1}
      {3}{{{\color{numb}3}}}{1}
      {4}{{{\color{numb}4}}}{1}
      {5}{{{\color{numb}5}}}{1}
      {6}{{{\color{numb}6}}}{1}
      {7}{{{\color{numb}7}}}{1}
      {8}{{{\color{numb}8}}}{1}
      {9}{{{\color{numb}9}}}{1}
}

\def\BibTeX{{\rm B\kern-.05em{\sc i\kern-.025em b}\kern-.08em
    T\kern-.1667em\lower.7ex\hbox{E}\kern-.125emX}}
\begin{document}

\title{Structured Extraction of Real World Medical Knowledge using LLMs for Summarization and Search\\
\thanks{This work is currently being funded by NIH contract 75N93024C00036.  Work performed by authors 1 and 2 contracted by Respond Health.}
}

\author{
\IEEEauthorblockN{
Edward Kim\textsuperscript{1,2}, 
Manil Shrestha\textsuperscript{1,2}, 
Richard Foty\textsuperscript{1}, 
Tom DeLay\textsuperscript{1}, 
Vicki Seyfert-Margolis\textsuperscript{1}
}
\IEEEauthorblockA{
\textsuperscript{1}RespondHealth, Washington, DC, USA \\
\textsuperscript{2}Department of Computer Science, Drexel University, Philadelphia, PA, USA 
}

\textit{\{ek826, ms5267\}@drexel.edu, \{rich.foty, tom.delay, vicki.seyfert-margolis\}@respondhealth.net}
}

\maketitle

%

\begin{abstract}
Creation and curation of knowledge graphs at scale can be used to exponentially accelerate the discovery, matching, and analysis of diseases in real-world data. While disease ontologies are useful for annotation, integration, and analysis of biological data, codified disease and procedure categories e.g. SNOMED-CT, ICD10, CPT, etc. rarely capture all of the nuances in a patient condition or, in the case of rare disease, may not even exist.  Furthermore, there are multiple disease definitions used in data sources and publications, each having its own structure and hierarchy. Mapping between ontologies, finding disease clusters, and building a representation of the chosen disease area are resource-intensive, often requiring significant human capital.  We propose the creation and curation of a patient knowledge graph utilizing large language model extraction techniques. In order to expand in volume and scale, knowledge graphs with generalized language capability allow for data to be extracted using natural language rather than being constrained by the exact terminology or hierarchy of existing ontologies. We develop a method of mapping back to existing ontologies such as MeSH, SNOMED-CT, RxNORM, HPO, etc. to ground the extracted entities to known entities in the medical community.  

We have access to one of the largest ambulatory care EHR databases in the country.  To demonstrate the effectiveness of our method, we benchmark our extraction in a test set with over 33.6M unique patients, in the area of patient search. In this case study, we perform a patient search for a rare disease: Dravet syndrome.  Dravet syndrome was codified as an ICD10 recognizable disease in October 2020.  In the following research, we describe our method of the construction of patient-specific knowledge graphs and subsequent searches for patients who exhibit symptoms of a particular disease.  Using patients with confirmed ICD10 codes for Dravet syndrome as our ground truth, we utilize our LLM-based entity extraction techniques and formalize an algorithmic way of characterizing patients in a grounded ontology to assist in mapping patients to specific diseases. Finally, we present the results of a real-world discovery method on Beta-propeller protein-associated neurodegeneration (BPAN), identifying patients with a rare disease, where no ground truth currently exists.
 \end{abstract}


\section{Introduction}
Knowledge graphs, as directed labeled graphs that model entities and their relationships, have emerged as a powerful paradigm for integrating heterogeneous data sources into a unified, semantically rich representation that enables complex reasoning and inference tasks across domains~\cite{ehrlinger2016towards, noy2019industry, hogan2020knowledge}.


\begin{figure}[tpb]
    \centering
    \includegraphics[width=0.9\linewidth]{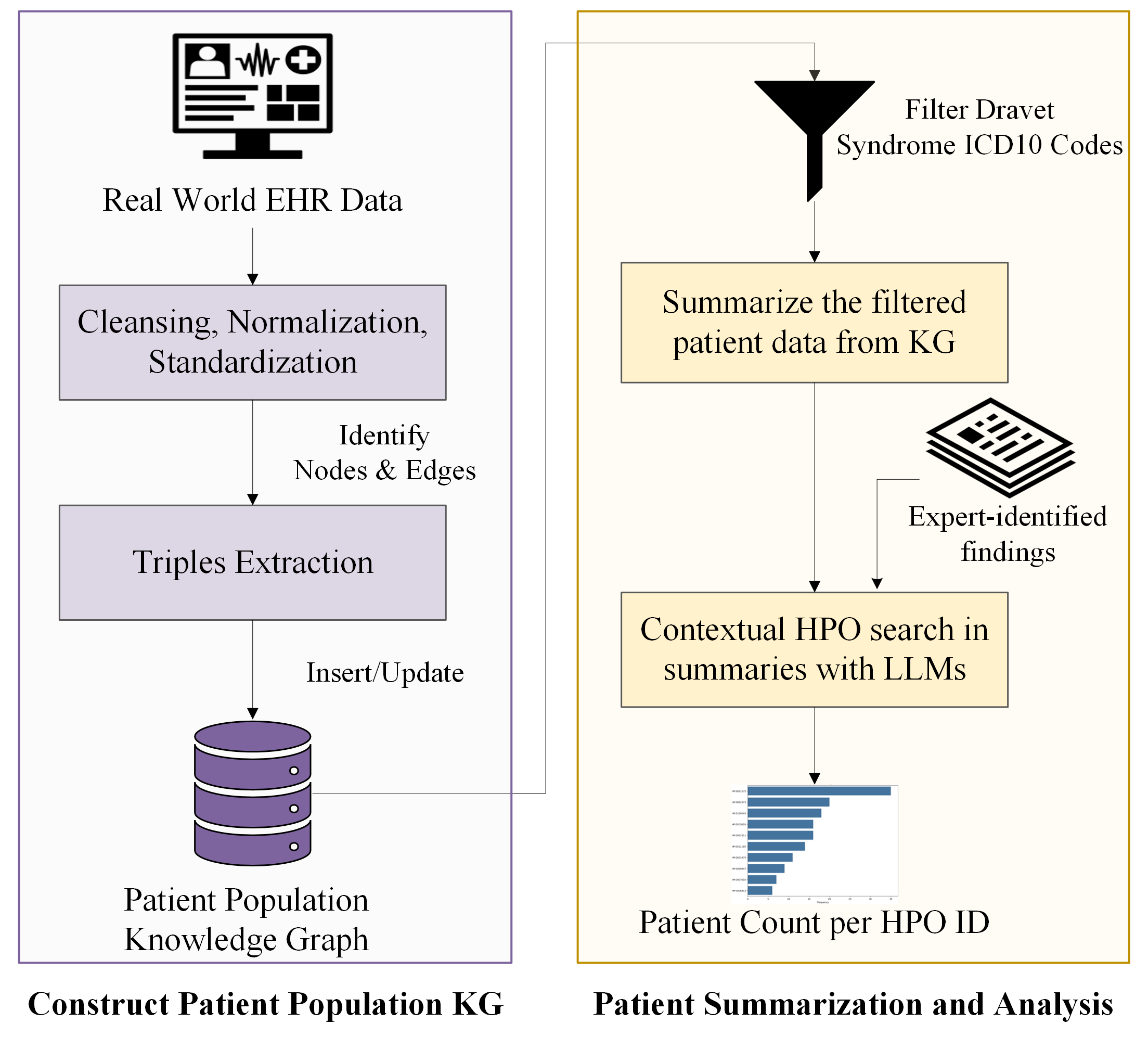}
    \caption{Visualization of knowledge graph construction and phenotype extraction for Dravet Syndrome patients. EHR data is structured into a knowledge graph with clinical entities (patients, symptoms, diagnoses, treatments) as nodes and their relationships as edges. Patient cohort is identified using Dravet Syndrome-specific ICD-10 codes (G40.83, G40.833, G40.834). Expert-curated Dravet Syndrome information is provided as context to the Large Language Model for HPO term extraction, as detailed in Section~\ref{realworld}.}
    \label{fig:front_page_graphics}
    \vspace{-1.2em} 
\end{figure}

Given the current and ever-expanding capabilities of the auto-regressive transformer architectures,  we hypothesize that Knowledge Graph (KG)-enabled Large Language Models (LLMs) will be ground-breaking for clinical research study design and development. The ability to extract data from published literature, clinical trial databases, anonymized patient medical records, and reported outcomes data will provide an enormous benefit in accelerating clinical study design. For example, analyzing patient inclusion criteria and endpoint measures across multiple disease studies could reveal which outcome measures proved most effective in specific clinical contexts. Importantly, this approach enables comparison between real-world effectiveness, derived from medical records, and reported outcomes in published literature across multiple studies of the same disease.

While AI can efficiently analyze vast amounts of scientific information, it is important to represent data in common data terminology and formats that enable linkable, shareable knowledge, which in turn, enable critical evaluation and reveal conceptual relationships. Knowledge graphs address these limitations by providing a framework for data representation, integration, and visualization that facilitates the identification of complex relationships across multiple data sources.
\\
Technical challenges in KG implementation include embeddings, acquisition, completion, fusion, and reasoning~\cite{ji2021survey}. The process of deriving knowledge from unstructured data, particularly medical text, is complicated by linguistic nuances and context-dependent meanings. Entity disambiguation presents additional complexity, especially when dealing with similar or identical names in different contexts. As highlighted in industry practices, successful semantic KG creation depends heavily on data harmonization across sources, though this is often hindered by inconsistent naming conventions and data heterogeneity across varying formats and standards~\cite{noy2019industry,hogan2020knowledge}.
\\
In this work, we propose the development of patient knowledge graphs and demonstrate the effectiveness of utilizing large language model extraction techniques. This approach enables data extraction via natural language, overcoming limitations of translation between medical notes to standardized ontologies. We present qualitative and quantitative results on several benchmark datasets as well as demonstrate a real-world case study on the mapping and discovery of two rare neurodegenerative diseases, Dravet Syndrome and BPAN.

\section{Background}
Knowledge graphs have become increasingly important in the medical domain, particularly for integrating heterogeneous data sources and enabling cross-domain knowledge discovery.
Traditionally, biomedical knowledge organizations relied heavily on manual curation by domain experts, a process that was time-consuming and expensive. Foundational resources include the Unified Medical Language System (UMLS)~\cite{bodenreider2004unified}, a comprehensive meta thesaurus of biomedical terminology, and Gene Ontology (GO)~\cite{ashburner2000gene}, which provides a structured vocabulary for gene and protein functions across species. Building upon these terminological frameworks, specialized biomedical knowledge graphs emerged, such as DisGeNET~\cite{pinero2016disgenet}, which focuses on gene-disease associations, and DrugBank~\cite{wishart2006drugbank,wishart2018drugbank,knox2024drugbank}, which captures comprehensive drug and drug target information including disease-drug-protein relationships. While these expert-curated resources provide reliable structured knowledge, their development requires significant financial resources and time-intensive manual curation. Moreover, despite their quality, they struggle to keep pace with the rapidly expanding medical literature.
\vspace{-0.3em} 
\subsection{Progress of Biomedical Knowledge Discovery}
The evolution of Natural Language Processing (NLP) in biomedical knowledge extraction can be characterized by three distinct eras. 

In the pre-transformer era (before 2018), traditional Named Entity Recognition (NER) relied heavily on rule-based systems and dictionary matching, with approaches like Conditional Random Fields (CRF) and Hidden Markov Models (HMM) being prominent for sequence labeling. Notable systems included MetaMap~\cite{aronson2001effective} for mapping biomedical text to UMLS concepts and DNorm~\cite{leaman2013dnorm} which introduced early deep learning for disease name normalization.

The next era would be characterized by the profound influence of transformer architectures~\cite{vaswani2017attention}. The introduction of BERT (Bidirectional Encoder Representations from Transformers) enabled a more nuanced understanding of the text~\cite{devlin2018bert}. In the biomedical domain, this led to specialized adaptations like BioBERT~\cite{lee2020biobert}, which demonstrated remarkable improvements in entity recognition tasks through additional pre-training on PubMed abstracts and PMC articles. Following this success, several domain-specific variants emerged: SciBERT~\cite{beltagy2019scibert} focused on scientific literature, ClinicalBERT~\cite{alsentzer2019publicly} specialized in clinical notes from electronic health records, and PubMedBERT~\cite{gu2021domain} took a unique approach by training exclusively on biomedical texts and publications. These specialized models significantly advanced our ability to extract and structure medical knowledge from text, particularly in tasks like biomedical entity recognition and relation extraction.

The emergence of LLMs post-2020 brought another paradigm shift. Domain-specific pretrained models like BioGPT~\cite{luo2022biogpt} and PhenoGPT \cite{yang2024enhancing} showed competitive performance, while general-purpose LLMs demonstrated impressive zero-shot and few-shot capabilities for entity recognition. Studies have shown LLMs outperforming fine-tuned BERT models on complex medical entity recognition tasks~\cite{agrawal2022large}, with efforts extending to automated medical KG construction from clinical notes~\cite{arsenyan2023large}.
\vspace{-0.3em} 
\subsection{Automated LLM Entity Extraction from Unstructured Data} Automation in knowledge graphs enables continuous data integration and enrichment from multiple sources, facilitating the incorporation of scientific data from research papers, patents, and clinical trials~\cite{yoo2020automating}. While manual validation remains essential, particularly for scientific accuracy, these automated systems significantly accelerate knowledge graph development. LLMs excel at text processing but suffer from hallucinations~\cite{maynez2020faithfulness}, whereas knowledge graphs provide structured, factual repositories. The integration of these technologies creates a synergistic system where LLMs enhance KG usability and process unstructured data, while knowledge graphs provide factual grounding~\cite{luo2024graph}. This combination results in AI systems that effectively merge language fluency with knowledge precision, addressing the limitations of each technology when used independently.
\vspace{-0.3em} 
\subsection{Enriching knowledge with known ontology}
Medical ontologies serve as standardized vocabularies for healthcare documentation and analysis, with several widely adopted standards including CPT~\cite{ama1966current},  ICD-10~\cite{world2004international}, SNOMED-CT~\cite{donnelly2006snomed}, HPO~\cite{robinson2008human}, and RxNORM~\cite{nelson2011normalized}. These ontologies gained prominence due to their comprehensive coverage and regulatory requirements: SNOMED-CT for clinical documentation, ICD-10 for disease classification and epidemiology tracking, CPT for procedure coding and reimbursement, RxNORM for medication standardization, and HPO for genetic disease classification. HPO has become particularly valuable in rare disease diagnosis workflows, where it helps connect clinical presentations to potential genetic causes.

The standardization and enriching of clinical notes with a known ontology is an important step in the ongoing challenge of patient documentation and the mapping of patient information~\cite{feng2022phenobert, yang2024enhancing, daniali2023enriching, moseley2020phenotype}. When properly implemented, these established ontologies enable a searchable knowledge base that improves research and healthcare system interoperability.
\vspace{-0.3em} 
\subsection{Background in Real World Datasets}
Through a collaborative partnership between RespondHealth (RH)~\cite{RespondHealth} and Harris Computer, RH’s test dataset represents data from approximately 33.6 million patients nationwide. This dataset integrates records from multiple EHR systems within the Harris Healthcare portfolio, covering diverse medical practices and specialties. The data includes extensive clinical documentation, with over 1 billion ICD-10 diagnoses, 6 billion lab values, nearly 470 million documented procedures, and unstructured clinical notes for each encounter. This comprehensive test dataset supports robust patient cohort identification and enables detailed analyses for research needs, from disease profiling to assessing treatment outcomes.  We will utilize this dataset to extract the phenotypic characteristics of patients and investigate their associations with clinical outcomes and treatment responses.

\section{Experiments}
In our work, we first describe and benchmark various techniques of structured entity extraction to specific ontologies.  We describe state-of-the-art prompting methodologies and fine-tuning methods to improve the precision and recall of LLM techniques and finally present experiments and results on real-world data.

\subsection{Experiment 1: Entity Extraction and Mapping to MeSH}
We first investigate the task of entity extraction. Entity extraction serves as a fundamental preprocessing step where we identify and classify textual elements that will form the basic components of the graph structure. For our evaluation, we utilize the BioCreative V Chemical Disease Relation (BC5CDR) dataset, a widely recognized benchmark in biomedical named entity recognition~\cite{li2016biocreative}. This corpus consists of PubMed abstracts manually annotated for chemical and disease entities. Our evaluation encompasses three distinct categories of models. The first category comprises fine-tuned BERT models, specifically BioBERT-Disease for disease entity recognition and BioBERT-Chemical for chemical entity extraction~\cite{lee2020biobert,dogan2014ncbi,alonso2021named}. The second category involves a traditional NER model, implemented as a spaCy NER model trained specifically on the BC5CDR corpus~\cite{neumann2019scispacy}. The third category explores LLMs through both zero-shot inference and few-shot learning implementations, as few-shot learning has shown promising results in various NLP tasks~\cite{brown2020language}, for which we use the Nemotron Llama 3.1-70B~\cite{wang2024helpsteer2preferencecomplementingratingspreferences} and Qwen2.5-72B models~\cite{qwen2.5}. We explore both static few-shot examples and Retrieval Augmented Generation (RAG) based dynamic few-shot examples, where few-shot examples are retrieved from the training set based on the cosine similarity between the embeddings of input text and the training text set. We utilize the gte-large-en-v1.5 model \cite{zhang2024mgte} for the embeddings throughout the experiments.  See Table \ref{tab:detailed_results} for detailed results and comparisons.

\begin{table}[tbh]
\centering
\footnotesize
\caption{Experiment 1: Biomedical Named Entity Recognition Performance: Traditional Models (BERT, SpaCy) versus LLMs in Zero-shot and Few-shot (FS) Settings}
\setlength\tabcolsep{3.0pt}
\begin{tabular}{llccc}
\hline
\multicolumn{1}{l}{\textbf{Model}} & \multicolumn{1}{l}{\textbf{Type}} & \multicolumn{1}{l}{\textbf{Precision}} & \multicolumn{1}{l}{\textbf{Recall}} & \multicolumn{1}{l}{\textbf{F1}} \\ \hline
\multicolumn{1}{l}{BioBERT} & Chemical & 0.838 & 0.800 & 0.818 \\
\multicolumn{1}{l}{BioBERT} & Disease & 0.765 & 0.820 & 0.791 \\ \hline
\multicolumn{1}{l}{SpaCy (en\_ner\_bc5cdr\_md)} & Chemical & 0.720 & 0.801 & 0.758 \\
\multicolumn{1}{l}{SpaCy (en\_ner\_bc5cdr\_md)} & Disease & 0.694 & 0.735 & 0.714 \\ \hline

\multicolumn{1}{l}{Llama3.1-Nemo-70B} & Chemical & 0.623 & 0.564 & 0.592 \\
\multicolumn{1}{l}{Llama3.1-Nemo-70B} & Disease & 0.710 & 0.507 & 0.592 \\ \hline
\multicolumn{1}{l}{Llama3.1-Nemo-70B w/ FS} & Chemical & 0.778 & 0.728 & 0.752 \\
\multicolumn{1}{l}{Llama3.1-Nemo-70B w/ FS} & Disease & 0.696 & 0.612 & 0.651 \\ \hline
\multicolumn{1}{l}{Llama3.1-Nemo-70B w/ Dyn. FS} & Chemical & 0.745 & 0.765 & 0.755 \\
\multicolumn{1}{l}{Llama3.1-Nemo-70B w/ Dyn. FS} & Disease & 0.685 & 0.640 & 0.662 \\ \hline

\multicolumn{1}{l}{Qwen2.5-72B} & Chemical & 0.555 & 0.658 & 0.602 \\
\multicolumn{1}{l}{Qwen2.5-72B} & Disease & 0.565 & 0.582 & 0.573 \\ \hline
\multicolumn{1}{l}{Qwen2.5-72B w/ FS} & Chemical & 0.671 & 0.786 & 0.724 \\
\multicolumn{1}{l}{Qwen2.5-72B w/ FS} & Disease & 0.590 & 0.686 & 0.634 \\ \hline
\multicolumn{1}{l}{Qwen2.5-72B w/ Dyn. FS} & Chemical & 0.617 & 0.801 & 0.697 \\
\multicolumn{1}{l}{Qwen2.5-72B w/ Dyn. FS} & Disease & 0.573 & 0.704 & 0.631 \\ \hline

\multicolumn{1}{l}{GPT-4o} & Chemical & 0.700 & 0.667 & 0.683 \\
\multicolumn{1}{l}{GPT-4o} & Disease & 0.710 & 0.560 & 0.626 \\ \hline
\multicolumn{1}{l}{GPT-4o w/ FS} & Chemical & 0.788 & 0.714 & 0.749 \\
\multicolumn{1}{l}{GPT-4o w/ FS} & Disease & 0.754 & 0.667 & 0.708 \\ \hline
\multicolumn{1}{l}{GPT-4o w/ Dyn. FS} & Chemical & 0.779 & 0.768 & 0.773 \\
\multicolumn{1}{l}{GPT-4o w/ Dyn. FS} & Disease & 0.768 & 0.696 & 0.730 \\ \hline
\end{tabular}
\label{tab:detailed_results}
\end{table}

\subsubsection{Lesson Learned: Training of Binary Encoders are Optimal}  In the case of a NER-based binary classification (is this a disease or not / is this a chemical or not), the most optimal performance can be achieved by training custom models for this task.  The BERT-based model is not only much smaller than the LLM-based decoder models, they are also orders of magnitude faster.

\subsubsection{Lesson Learned: Dynamic Few Shot Prompting Yields Optimal Performance for LLMs} While the LLM (decoder architectures) are nearly on par with the trained encoder models, they benefit significantly from in-context learning using few-shot prompting and dynamic few-shot techniques.  The dynamic few-shot retrieves the top 5 most similar abstracts via cosine similarity of the text embeddings.  Even though the smaller models outperform the LLMs in this binary task, we note that dynamic few-shot prompting is a much more flexible and generalizable approach.

\subsection{Experiment 2: Multi-label Phenotype Classification}
For the second experiment, we investigate the benchmark performance of LLMs in multi-label multi-class phenotype classification tasks. This experiment evaluates a significantly more complex task than the previous experiment that requires understanding the entire context of medical notes and making multiple, potentially interrelated decisions about patient conditions. 

For our evaluation, we utilize the phenotype annotations dataset for patient EHR notes in the MIMIC-III database \cite{moseley2020phenotype,goldberger2000physiobank,johnson2016mimic,gehrmann2018comparing}. There are 844 distinct patient notes in this dataset which have been annotated by two annotators for 15 different categories: 13 phenotypes, plus categories for None and Unsure. We use 422 notes as our test set across all experimental settings, with the remaining notes serving as reference examples specifically for the dynamic RAG-based few-shot setting. 
We selected two families of models to evaluate in this experiment: the encoder-only BERT model as our baseline and decoder-only LLMs. The chosen BERT model is the DeBERTa-v3-large model fine-tuned for Natural Language Inference presented by \cite{laurer2024less}, which can be used for zero-shot classification. Similar to Experiment 1, we evaluate LLMs in three context settings: zero-shot, static few-shot, and dynamic RAG-based few-shot.  See Table \ref{tab:model-comparison} for detailed results and comparisons.

\begin{table}[tbh]
\centering
\footnotesize
\caption{Experiment 2: Multilabel-Multiclass Phenotype Classification Performance on MIMIC-III EHR Notes}
\setlength\tabcolsep{2.5pt}
\begin{tabular}{lcccc}
\hline
\textbf{Model} & \textbf{Precision} & \textbf{Recall} & \textbf{F1} & \textbf{Micro Acc.} \\
\hline
Finetuned DeBERTa-v3-large \cite{laurer2024less} & 0.489 & 0.591 & 0.455 & 0.799 \\\hline
Qwen2.5-72B & 0.645 & 0.737 & 0.653 & 0.888 \\
Qwen2.5-72B w/ FS & 0.546 & 0.733 & 0.579 & 0.855 \\
Qwen2.5-72B w/ Dyn. FS & 0.730 & 0.865 & 0.761 & 0.927 \\\hline
Llama3.1-Nemo-70B & 0.679 & 0.644 & 0.609 & 0.892 \\
Llama3.1-Nemo-70B w/ FS & 0.649 & 0.643 & 0.562 & 0.873 \\
Llama3.1-Nemo-70B w/ Dyn. FS & 0.747 & 0.804 & 0.732 & 0.919 \\ \hline
\multicolumn{5}{l}{*Can not run it on GPT-4o due restrictions as per Physionet DUA \cite{PhysioNet_GPT_Responsible_Use}} \\\hline
\end{tabular}
\label{tab:model-comparison}
\end{table}
\vspace{-0.1em}
\subsubsection{Lesson Learned: In the Multi-Class Categorization of Medical notes, the Deep Understanding Capabilities in an LLM are needed}  In the first experiment, when it was simply a binary classification of a particular word, the BERT encoder models were sufficient and optimized; however, in the case of a complex, multi-class classification setting, one needs to utilize the deep understanding capabilities of a large language model.  We also note that the few shot prompting techniques lessons learned also hold in this experiment.
\vspace{-0.3em} 
\subsection{Experiment 3: Extracting and Mapping HPO terms from Clinical Notes}
Experiment 3 is a hybrid of the previous two tasks.  In this experiment, we are both performing a medical NER, but then doing a classification of that entity extracted to a specific Human Phenotype Ontology (HPO \cite{robinson2008human}) ID that contains the preferred name, synonyms, and definition.  

The HPO  is a standardized vocabulary for describing human disease manifestations and phenotypic abnormalities. It provides a comprehensive, hierarchically organized database of human phenotypes (observable characteristics) that enables precise descriptions of clinical features in both rare and common diseases. The HPO was developed to support computational analysis of human disease, allowing researchers and clinicians to describe patient symptoms in a standardized way, facilitate diagnosis, and connect similar cases across different medical centers. It contains over 13,000 terms and serves as a crucial resource in clinical diagnostics, research, and development.

For this experiment, we benchmark against the BiolarkGSC+ dataset \cite{yan2022phenorerank}. The BiolarkGSC+ dataset is an updated version of the GSC dataset, Bio-Lark Gold Standard Corpus, that consists of 228 de-identified clinical notes with labeled HPO terms.  We re-implemented the method from PhenoGPT \cite{yang2024enhancing} to build the GSC+ validation set, that consists of 23 test notes and 200 training notes. In our LLM fine-tuned models, we trained the Llama3.1-Nemo-70B model using supervised fine-tuning with a QLoRA for the training of a 64-rank attention adapter.  The LoRA was fine-tuned for 2 epochs over the training set (more epochs hurt performance).  

In Table \ref{tab:hpo-comparison} we present some previous work in the rule-based HPO extraction and mapping \cite{liu2019doc2hpo,deisseroth2019clinphen}, deep learning BERT-based implementations \cite{luo2021phenotagger,yan2022phenorerank} and LLM based methods \cite{yang2024enhancing}.  The task requires a correct extraction of entities, and correct mapping of that entity to a specific HPO ID. 

\begin{table}[tbh]
\centering
\footnotesize
\caption{Experiment 3: Performance of various methods on the BiolarkGSC+ (validation) dataset involving HPO extraction and mapping.}
\setlength\tabcolsep{2.5pt}
\begin{tabular}{lccc}
\hline
\textbf{Model} & \textbf{Precision} & \textbf{Recall} & \textbf{F1} \\
\hline
Doc2hpo-Ensemble \cite{liu2019doc2hpo} & 0.754 & 0.608 & 0.673 \\
ClinPhen \cite{deisseroth2019clinphen} & 0.590 & 0.418 & 0.489 \\
Phenotagger \cite{luo2021phenotagger} & 0.720 & 0.760 & 0.740 \\
PhenoReRank \cite{yan2022phenorerank} & 0.843 & 0.708 & 0.770 \\
\hline
Qwen2.5-72B & 0.747 & 0.483 & 0.587 \\
Qwen w/ Glean & 0.760 & 0.535 & 0.628 \\
Qwen2.5-72B w/ FS & 0.734 & 0.542 & 0.623 \\
Qwen2.5-72B w/ FS\&Glean & 0.732 & 0.565 & 0.638 \\
Llama3.1-Nemo-70B & 0.614 & 0.464 & 0.528 \\
Llama3.1-Nemo-70B w/ Glean & 0.681 & 0.528 & 0.595 \\
Llama3.1-Nemo-70B w/ FS & 0.729 & 0.524 & 0.609 \\
Llama3.1-Nemo-70B w/ FS\&Glean & 0.667 & 0.599 & 0.631 \\

GPT-4o & 0.788 & 0.476 & 0.594 \\
GPT-4o w/ Glean & 0.762 & 0.516 & 0.616 \\
GPT-4o w/ FS & 0.738 & 0.545 & 0.627 \\
GPT-4o w/ FS\&Glean & 0.765 & 0.597 & 0.671 \\\hline
Llama3.1-Nemo-70B LoRA (PhenoGPT \cite{yang2024enhancing})& 0.753 & 0.617 & 0.678 \\
Llama3.1-Nemo-70B LoRA w/ Glean & 0.775 & 0.673 & 0.720 \\
\hline
\end{tabular}
\label{tab:hpo-comparison}
\end{table}

\subsubsection{Lesson Learned: Finetuning BERT-based models or LLM models for Specific Ontologies Yields Optimal Results}  In this experiment, we note that optimal performance can be achieved by fine-tuned models.  While the BERT-based models edge out the LLMs, as this is a NER-based experiment, fine-tuning an LLM for this specific task puts the LLM on par with encoder models.  In fact, in the PhenoGPT \cite{yang2024enhancing} paper, the LLM-based models were documented to outperform the BERT-based models on this same dataset.  While re-implementing their method, which showed improvement over our base model, we were not able to get the same improvement boost in our model.  

\subsubsection{Lesson Learned: Narrowly Finetuning LLM models Degrades Generalization}  While using a LoRA noted improvement over this narrow task, we remain skeptical about this methodology in the real world.  In fact, we note that this dataset only contains 184 unique phenotypes in an HPO set of over 13,000, and the narrow fine-tuning of the LLM to boost performance on the BiolarkGSC+ benchmark is not generalizable to diverse notes with an evolving set of HPO terms.  Thus, we believe that experimental results may not translate to the real world.

\subsubsection{Lesson Learned: Gleaning improves Recall}  The process of gleaning with LLMs involves taking the results of the previous extraction and redoing the extraction process over multiple steps \cite{edge2024local}.  The previous extraction is entered back into the context window and the model is prompted to both look at what was extracted previously and find new entities to extract.  We noticed that in this particular dataset, we were seeing fairly low recall; many of the HPO entities were not being extracted.  When gleaning for a single iteration, we are able to significantly increase recall, while mostly also improving precision (or keeping the performance of precision relatively flat).

\section{Real World Case Study and Results}
\label{realworld}
\begin{figure*}[tpb]
    \centering
    \includegraphics[width=0.7\linewidth]{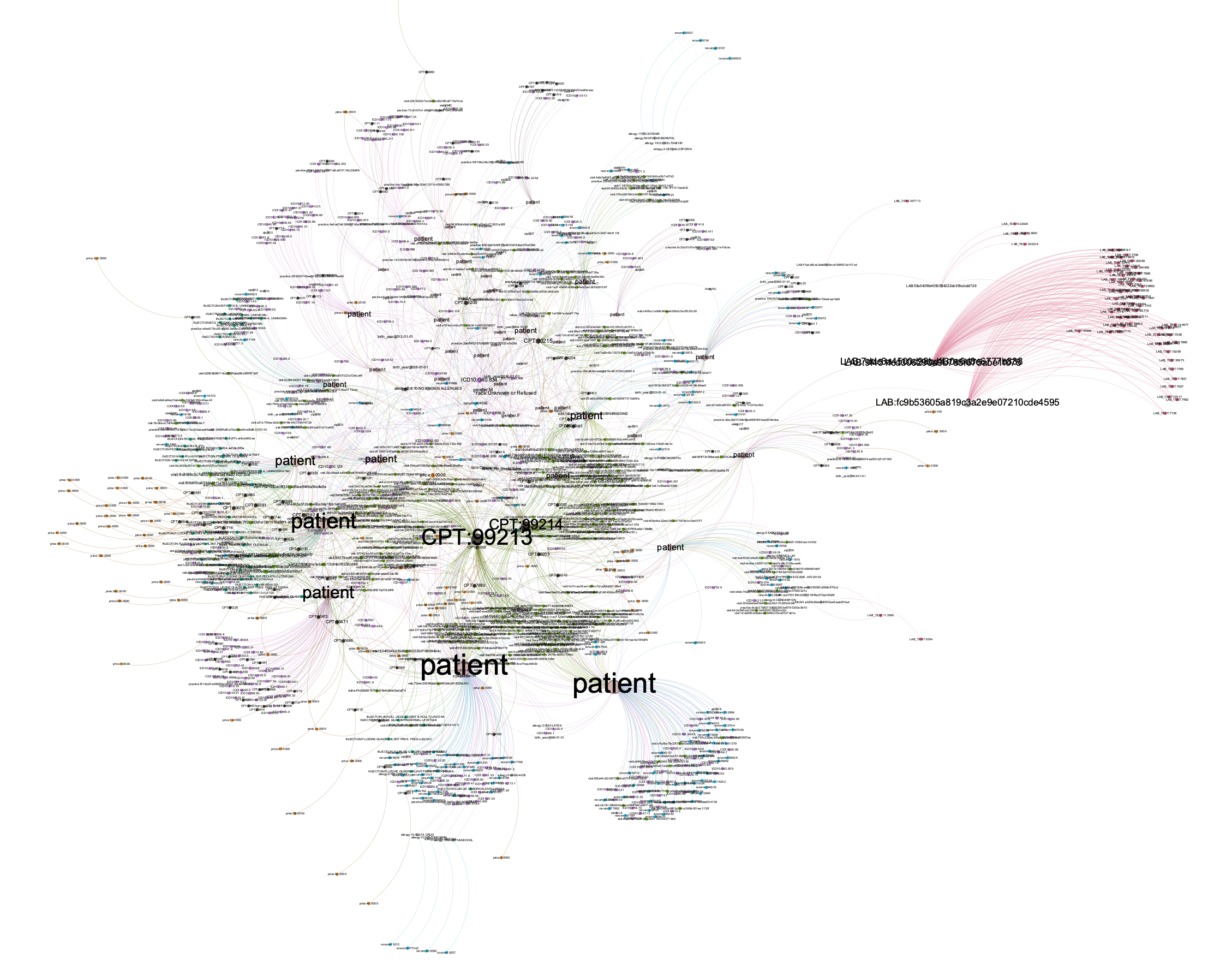}
    \caption{Visualization of the knowledge graph of 38 ICD-10 coded patients with Dravet.  The knowledge graph consists of both structured elements from the EHR, as well as a dearth of unstructured information in the medical notes, and other multimodal forms.  Patients and their medical information are mapped via unique patient keys and the unstructured data resides as node attributes.  Our case study involves the construction of KGs as well as the extraction of HPO-related terms from the unstructured data.}
    \label{fig:patients}
    \vspace{-0.3em}
\end{figure*}

Using all of the lessons learned in previous experimentation, we performed patient mapping and search over our real-world dataset.  As mentioned before, our test dataset consists of mostly ambulatory data covering 33.6M patients.  We focus on two rare diseases: Dravet Syndrome and BPAN.  
\vspace{-0.4em} 
\subsection{Description of the Rare Diseases} 
\textit{Dravet Syndrome} - Dravet syndrome is a severe form of epilepsy that manifests during the first year of life in otherwise healthy infants. The condition, primarily caused by mutations in the SCN1A gene affecting sodium ion channels in the brain, represents a complex neurological disorder that impacts multiple aspects of development and daily functioning.

The syndrome typically announces itself through prolonged seizures, often triggered by elevated body temperature. As the condition progresses, children develop multiple types of seizures before age five, including particularly concerning episodes of status epilepticus that require immediate medical intervention. While patients initially present with normal EEG readings and development, significant changes emerge during the second and third years of life, when both EEG abnormalities and developmental delays become apparent.

The impact of Dravet syndrome extends far beyond seizures. Patients commonly experience significant developmental challenges, with most cases resulting in moderate to severe delays. Speech impairment often becomes noticeable before age two, and movement difficulties, characterized by poor coordination (ataxia) and low muscle tone (hypotonia), persist throughout life. These physical challenges may worsen over time, potentially leading to decreased mobility during adolescence. The condition's complexity is further compounded by sleep disturbances, behavioral issues, and disruptions to the autonomic nervous system, affecting basic functions like temperature regulation and sweating.

The prognosis for individuals with Dravet syndrome typically includes long-term dependence on caregivers, though the severity can vary. Management of the condition requires a comprehensive approach, incorporating physical, occupational, and speech therapy to address the multiple challenges these patients face. Despite these interventions, Dravet syndrome remains a significant medical challenge that substantially impacts both patients and their families throughout their lives.

\textit{Beta-propeller protein-associated neurodegeneration} - 
Beta-propeller protein-associated neurodegeneration (BPAN) is a progressive neurological disorder characterized by the gradual accumulation of iron in the brain. This condition, classified as a type of neurodegeneration with brain iron accumulation (NBIA), presents a complex array of symptoms that evolve and worsen over time. The disorder typically manifests in infancy or early childhood with various types of seizures. These can range from febrile seizures triggered by high temperatures to more severe generalized tonic-clonic seizures that affect the entire body. Patients may experience multiple types of seizures, including absence seizures resembling daydreaming spells, atonic seizures characterized by sudden muscle weakness, myoclonic seizures involving muscle twitches, and epileptic spasms. Some cases present seizure patterns similar to those seen in West syndrome or Lennox-Gastaut syndrome.

During childhood, BPAN patients struggle with significant developmental challenges, including intellectual disability and pronounced difficulties with expressive language and motor coordination. The condition often presents features reminiscent of Rett syndrome, such as stereotypic hand movements, teeth grinding, sleep disturbances, and autism-like characteristics affecting communication and social interaction. Movement difficulties (ataxia) impact both gross motor skills like walking and fine motor skills such as using utensils.

A significant turning point in the disease occurs during late adolescence or early adulthood when patients begin experiencing cognitive decline that can progress to severe dementia. This period also marks the onset of worsening movement disorders, including dystonia (particularly affecting the arms) and parkinsonism. The Parkinsonian symptoms include slow movement, rigidity, tremors, postural instability, and a distinctive shuffling gait that increases the risk of falls.

While individuals with BPAN can live into middle age with proper medical management, the condition ultimately proves fatal, often due to complications from dementia or movement-related problems such as fall injuries or aspiration pneumonia resulting from swallowing difficulties. This progressive disorder presents significant challenges for both patients and caregivers, requiring comprehensive medical care and support throughout the patient's life.

\subsection{Patient Knowledge Graph Construction for Dravet Syndrome Cases}
One of the reasons for studying Dravet Syndrome is the recent introduction of this rare disease in the ICD-10 coding scheme.  The new designation G40.83 and its subcategories provide precise coding options for Dravet syndrome, also known as Polymorphic epilepsy in infancy (PMEI) or Severe myoclonic epilepsy in infancy (SMEI). The codes further differentiate between cases that are intractable with status epilepticus (G40.833) and those that are intractable without status epilepticus (G40.834).

In our dataset, we have a total of 38 unique patients coded under G40.83, G40.833, and G40.834.  For all of these patients, our EHR record contains, both structured data (ICD-10 coding, CPT coding, medications to RxNorm, demographic data like age, race, state, and zip) and unstructured data (medical notes, previous medical history, current visit purpose, imaging, genetics data in PDF form, etc.).  Using a combination of structuring techniques we can build a preliminary KG of the patients and their attributes as visualized in Figure \ref{fig:patients}.  

For BPAN, we do not have ground truth data.  From examining our study of Dravet syndrome, patients were often categorized under broader codes such as G40.8 (Other epilepsy and recurrent seizures), which failed to capture the complex nature and specific healthcare needs of the condition. This generic classification posed several challenges for the patient community, including difficulties in securing appropriate medical coverage and accessing necessary treatments for the various comorbidities associated with the syndrome.  BPAN is currently in this stage as there is no ICD-10 designation for this disease, thus the disease does fall within more generic codes and likely remains underdiagnosed and lacks recognition in the medical community.
\vspace{-0.3em} 
\subsection{HPO classification of our Patient Knowledge Graph}  The HPO provides the phenotypical presentations of diseases, as well as their frequency of presentation in a typical patient.  For Dravet syndrome there are 48 identified phenotypes as shown in Table \ref{tab:hpo-types}.  We utilize LLMs to view the patient knowledge graph, and for each patient, we perform a multi-class categorization of their entire patient history and identify which of these HPO entities are present.  The main extraction goals are the following, 
\begin{lstlisting}[language=JSON]
Your goal is to extract any of the provided HPOs given patient details. Please present the output in the following JSON format.
Result:{
    "<PATIENT_KEY>": [
        {"category": "HPO Identifier e.g. HP:0100694",
            "confidence": "confidence score between 0 and 1",
            "reasoning": "Why it qualifies as this HPO"
        }]
    }
\end{lstlisting}
The full prompt can be seen in the Appendix.  

\subsection{Frequency in the Real World}
\begin{figure*}[tpb]
    \centering
    \includegraphics[width=0.81\linewidth]{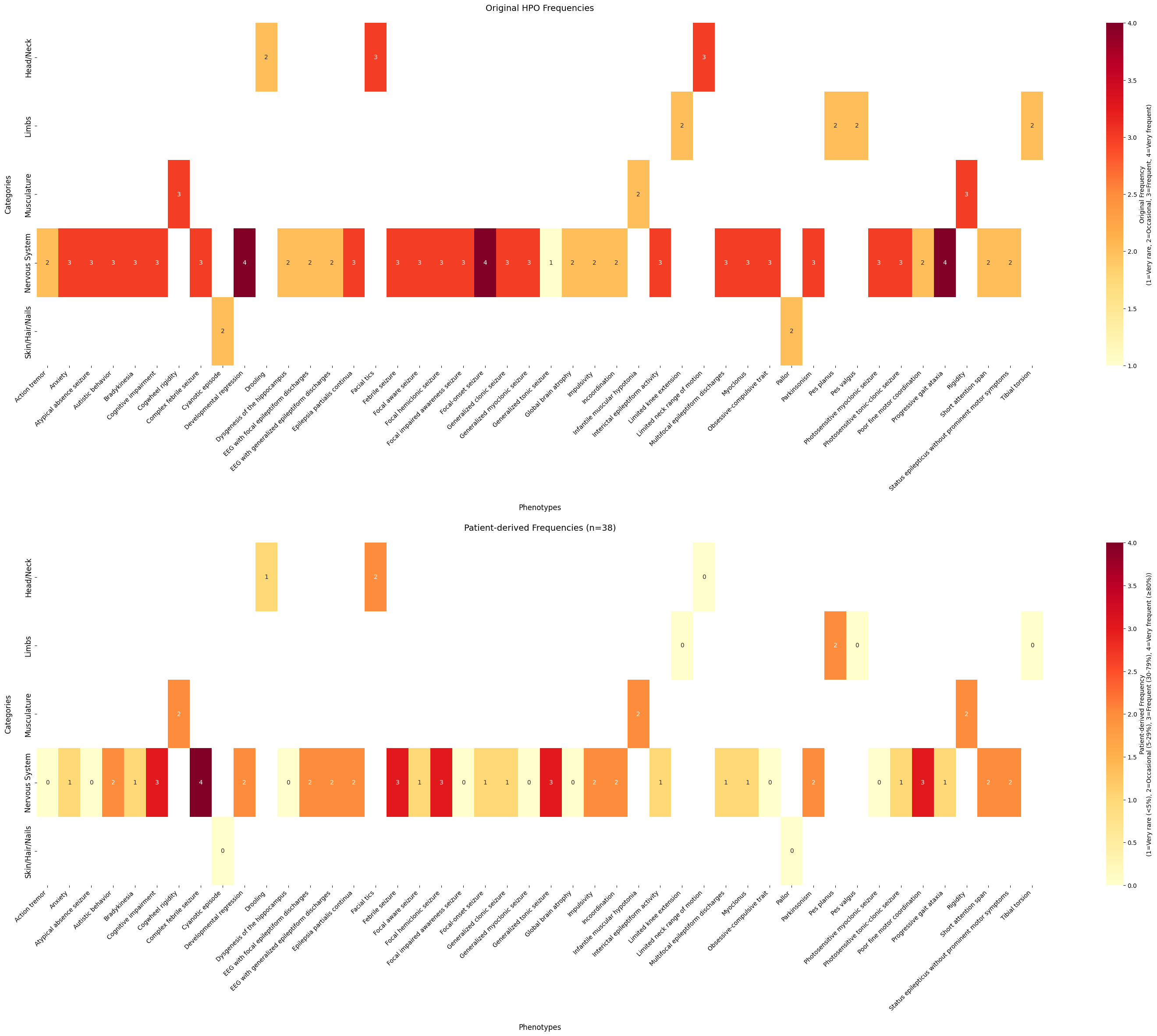}
    \caption{Heat map of frequencies of presentation in the HPO and frequencies present in real-world Dravet syndrome patients.  Comparison between the two enables the discovery of clinical presentation in a diverse, real-world dataset.  The largest differences are in the head and neck/limb presentation, where many of the phenotypes are not mentioned within the clinical notes of confirmed Dravet patients.}
    \label{fig:freq}
\end{figure*}

In the HPO, frequency refers to how often a particular phenotypic feature (symptom or characteristic) occurs in individuals with a specific disease or genetic condition. The HPO uses standardized frequency categories as follows, Very rare (1-4\%), Occasional (5-29\%), Frequent (30-79\%), Very frequent (80-99\%), and Obligate (100\%).  Our extraction methodology classifies the presentation of these phenotypes in our patient knowledge graph and can be seen in Table \ref{tab:hpo-types}, and a detailed comparison of their frequencies in the real-world data as compared to the frequencies noted by the HPO can be seen in Figure \ref{fig:freq}.  Some of the notable differences are in the absence of (or at least the absence of noted phenotypes in the clinical notes) symptomatic presentation in the head, neck, and limbs along with a few nervous system phenotypes.  Only three phenotypes were more frequent than noted in the HPO, febrile seizures, generalized tonic seizures, and poor fine motor coordination.

\begin{table}
\centering
\caption{Patient count per HPO ID in 38 Dravet Syndrome cases, extracted using Llama3.1-Nemo-70B}
\begin{tabular}{llc}
\hline
\textbf{HPO ID} & \textbf{Description} & \textbf{Frequency} \\
\hline
HP:0011172 & Complex febrile seizure & 34 \\
HP:0002373 & Febrile seizure (age 3 mo. to 6 yrs.)  & 24 \\
HP:0100543 & Cognitive impairment  &  23 \\
HP:0006813 & Focal hemiclonic seizure    &  16 \\
HP:0010818 & Generalized tonic seizure  &   13 \\
HP:0007010 & Poor fine motor coordination   &   12 \\
HP:0011185 & EEG with focal epileptiform discharges  &  11 \\
HP:0008947 & Infantile muscular hypotonia   &   9 \\
HP:0011198 & EEG with generalized epileptiform dis.   &   8 \\
HP:0000729 & Autistic behavior   &  6 \\
HP:0002376 & Developmental regression    &  6 \\
HP:0000736 & Short attention span   &   4 \\
HP:0012847 & Epilepsia partialis continua   &   4 \\
HP:0100710 & Impulsivity  & 4 \\
HP:0001763 & Pes planus   & 3 \\
HP:0011468 & Facial tics  & 3 \\
HP:0031475 & Status epilepticus w/o prominent motor &  3 \\
HP:0001300 & Parkinsonism    &  2 \\
HP:0002063 & Rigidity   &   2 \\
HP:0002311 & Incoordination  &  2 \\
HP:0002396 & Cogwheel rigidity    & 2 \\
HP:0000739 & Anxiety  & 1 \\
HP:0001336 & Myoclonus   &  1 \\
HP:0002067 & Bradykinesia    &  1 \\
HP:0002307 & Drooling    &  1 \\
HP:0002349 & Focal aware seizure  &  1 \\
HP:0007207 & Photosensitive tonic-clonic seizure  & 1 \\
HP:0007240 & Progressive gait ataxia  & 1 \\
HP:0007359 & Focal-onset seizure  & 1 \\
HP:0010841 & Multifocal epileptiform discharges  &  1 \\
HP:0011169 & Generalized clonic seizure   & 1 \\
HP:0011182 & Interictal epileptiform activity   &   1 \\
HP:0000466 & Limited neck range of motion & - \\
HP:0000980 & Pallor & - \\
HP:0001327 & Photosensitive myoclonic seizure & - \\
HP:0002123 & Generalized myoclonic seizure & - \\
HP:0002283 & Global brain atrophy & - \\
HP:0002345 & Action tremor & - \\
HP:0002384 & Focal impaired awareness seizure & - \\
HP:0003066 & Limited knee extension & - \\
HP:0007270 & Atypical absence seizure & - \\
HP:0008081 & Pes valgus & - \\
HP:0008770 & Obsessive-compulsive trait & - \\
HP:0025101 & Dysgenesis of the hippocampus & - \\
HP:0100694 & Tibial torsion & - \\
HP:0200048 & Cyanotic episode & - \\
\hline
\end{tabular}
\label{tab:hpo-types}
\end{table}
\vspace{-0.2em}
\section{Discussion and Future Work}
\vspace{-0.1em}
In the real-world case of Dravet syndrome, we have ICD-10 coded patients; however, as we begin to explore more rare diseases, oftentimes there is no code that enables us to discover and analyze particular patient cohorts.  In our second case study on BPAN, we fall into this category of rare disease where no ICD-10 codes exist, and the difficulty is in finding these ``needle-in-a-haystack'' patients.  If we can identify disease via their phenotypic presentations, we may be able to help patients by connecting them with resources, literature, and educational materials.

From a keyword search for ``BPAN'', we were able to identify 2 confirmed patients via their unstructured clinical notes and analyze them further.  There were six generic ICD-10s that the two exhibited were R62.50 (Unspecified lack of expected normal physiological development in childhood), G40.219 (Epilepsy, unspecified, not tractable, without status epilepticus), G23.8 (Other specified degenerative diseases of basal ganglia), F79 (unspecified intellectual disabilities), G40.824 (epilepsy with centrotemporal spikes), and G31.9 (nonspecific code for degenerative disease).  As one can see from this list, each of these codes is very generic.  A total search of all patients matching one of these codes is over 24k patients.  We then utilize a proprietary method to rate each patient on a scale of 0-9 on their likelihood of BPAN.  268 patient records were in the 7-9 range, then each of these patients was extracted into a knowledge graph and analyzed for phenotypic presentations of BPAN.  We were able to finally narrow this patient list down to 12 very high probability cases of BPAN that remain undiagnosed.  At this time, we are exploring multiple possible avenues on how to utilize our findings in order to inform and help these patients with careful consideration of how this information could impact their livelihood.

The framework could be readily adapted to other rare diseases lacking specific diagnostic codes, leveraging both the language model's flexibility with clinical descriptions and the adaptable knowledge graph structure.
\vspace{-0.3em}
\section{Conclusion}
Our analysis demonstrates the effectiveness of leveraging large language models for structured extraction and patient-specific knowledge graph construction in rare disease identification.  We experimented with several off-the-shelf benchmark datasets, quantified our methods and results, and presented lessons learned in order to translate this for real-world data.  In our real-world dataset, by comparing standardized HPO frequencies against observed frequencies in a cohort of 38 Dravet Syndrome cases, we revealed both the strengths and limitations of current ontological frameworks. The frequency discrepancies between established HPO annotations and real-world patient data highlight the need for dynamic, data-driven approaches to disease characterization. Our method of using LLM-based extraction techniques, combined with mapping to established ontologies like HPO, provides a scalable solution for bridging the gap between standardized disease definitions and the complex reality of patient presentations. This approach proves particularly valuable for rare diseases like Dravet Syndrome and BPAN, where traditional coding systems may be insufficient or nonexistent. 

\section{Acknowledgments}
This work is currently being funded by NIH contract 75N93024C00036.  Thanks to Sammi Mentis for project management, and Ben Siegler for his exploratory work in effectively interfacing with KGs using LLMs.
\bibliographystyle{unsrt}
\footnotesize
\bibliography{references}
\appendix
\begin{figure*}[ht!]
    \begin{lstlisting}[caption={Prompt for HPO Phenotype Extraction Experiment},
    label={lst:hpo-prompt}]
You are an expert in mapping clinical phenotypes to the Human Phenotype Ontology (HPO). Your task is to analyze provided patient information and clinical descriptions, identifying all relevant phenotypic abnormalities and clinical features present in the text. For each identified feature, you should match it to the most appropriate HPO term and its corresponding ID.

INPUT FORMAT: Patient information will be provided as as structured entry. Each patient may contain multiple observations, symptoms, or conditions and extra information.

RESPONSE FORMAT:
- MUST be a single JSON object
- NO explanatory text, notes, or comments
- NO markdown formatting
- NO additional fields beyond the specified format

Dravet syndrome (DS) is a serious form of epilepsy that appears in otherwise healthy infants during their first year. It often begins with prolonged seizures lasting over five minutes, typically generalized tonic-clonic or hemiclonic. As children grow, they tend to experience additional seizure types before age five, with many seizures triggered by fevers.
The majority of Dravet syndrome cases, over 80%, are linked to mutations in the SCN1A gene, which affects sodium ion channels that are vital for electrical signaling in the brain and heart. This dual impact classifies DS as both an epileptic encephalopathy and a channelopathy, influencing brain function through seizure activity and channel dysfunction.
While infants with Dravet syndrome usually develop normally at first, developmental delays often become evident in the second and third years. The condition poses various challenges, including speech difficulties, poor coordination (ataxia), and low muscle tone (hypotonia). Sleep issues and behavioral problems are common, along with disruptions to the autonomic nervous system, which can affect basic functions like temperature control.
The long-term prognosis for individuals with Dravet syndrome can vary significantly, but most require caregiver support into adulthood. Mobility often declines during adolescence, and moderate to severe developmental delays are typical. Treatment generally involves emergency measures for prolonged seizures, as well as extensive physical, occupational, and speech therapy. Additionally, care needs to address other health concerns, such as growth and nutrition issues.

Here are the HPO phenotypes for Dravets:
HP:0002345 	 Action tremor
HP:0000739 	 Anxiety
HP:0007270 	 Atypical absence seizure
HP:0000729 	 Autistic behavior
HP:0002067 	 Bradykinesia
<Comment: There are a total of 46 HPOs associated with DS, removed them for space here>
HP:0007010 	 Poor fine motor coordination
HP:0007240 	 Progressive gait ataxia
HP:0002063 	 Rigidity
HP:0000736 	 Short attention span
HP:0031475 	 Status epilepticus without prominent motor symptoms
HP:0100694 	 Tibial torsion

GOAL: Your goal is to extract any of the provided HPOs given patient details. Please present the output in the following JSON format.

Result:{
    "<PATIENT_KEY>": [
        {
            "category": "HPO Identifier e.g. HP:0100694",
            "confidence": confidence score between 0 and 1
            "reasoning": Why do you think it qualifies as this HPO
        }
    ]
}

REMEMBER RESPONSE FORMAT:
- MUST be a single JSON object
- NO explanatory text, notes, or comments
- NO markdown formatting
- NO additional fields beyond the specified format

NEVER include extra fields or explanations. ANY deviation from this format is an error.
    \end{lstlisting}
\end{figure*}

\end{document}